\documentclass{article}

\usepackage{microtype}
\usepackage{graphicx}
\usepackage{subcaption}
\usepackage{booktabs} 
\usepackage{amsmath}
\usepackage{amssymb}
\usepackage{mathtools}
\usepackage{algorithm}
\usepackage{algorithmic}
\usepackage{adjustbox}
\newcommand{\RETURN}{\textbf{return} }
\usepackage{tabularx}
\usepackage[table,dvipsnames]{xcolor} 
\usepackage{multirow}
\usepackage{enumitem}
\usepackage{booktabs}
\usepackage{array}
\usepackage{svg}
\usepackage{hyperref}
\usepackage[capitalize,noabbrev]{cleveref}
\crefname{appendixsection}{Appendix}{Appendices}
\DeclareMathOperator*{\argmin}{argmin}

\usepackage[preprint]{icml2026}

\icmltitlerunning{SCE-LITE-HQ: High-Resolution Counterfactual Explanations}

\begin{document}

\title{SCE-LITE-HQ: Smooth visual counterfactual explanations with generative foundation models}

\twocolumn[
  \icmltitle{SCE-LITE-HQ: Smooth visual counterfactual explanations with generative foundation models}



  \icmlsetsymbol{equal}{*}

  \begin{icmlauthorlist}
    \icmlauthor{Ahmed  Zeid}{yyy}
    \icmlauthor{Sidney Bender}{yyy,zzz}
  \end{icmlauthorlist}

 \icmlaffiliation{yyy}{Machine Learning Group, Technische Universität Berlin, Berlin, Germany}
  \icmlaffiliation{zzz}{BIFOLD – Berlin Institute for the Foundations of Learning and Data, Berlin, Germany}

  \icmlcorrespondingauthor{Ahmed  Zeid}{a.zeid@tu-berlin.de}
  \icmlcorrespondingauthor{Sidney Bender}{s.bender@tu-berlin.de}

  \icmlkeywords{Machine Learning, ICML}

  \vskip 0.3in
]



\printAffiliationsAndNotice{}  

\begin{abstract}
Modern neural networks achieve strong performance but remain difficult to interpret in high-dimensional visual domains. Counterfactual explanations (CFEs) provide a principled approach to interpreting black-box predictions by identifying minimal input changes that alter model outputs. However, existing CFE methods often rely on dataset-specific generative models and incur substantial computational cost, limiting their scalability to high-resolution data.
We propose SCE-LITE-HQ, a scalable framework for counterfactual generation that leverages pretrained generative foundation models without task-specific retraining. The method operates in the latent space of the generator, incorporates smoothed gradients to improve optimization stability, and applies mask-based diversification to promote realistic and structurally diverse counterfactuals. We evaluate SCE-LITE-HQ on natural and medical datasets using a desiderata-driven evaluation protocol. Results show that SCE-LITE-HQ produces valid, realistic, and diverse counterfactuals competitive with or outperforming existing baselines, while avoiding the overhead of training dedicated generative models.

\end{abstract}


\begin{figure*}[t!]
    \centering
    \includegraphics[width=0.9\linewidth]{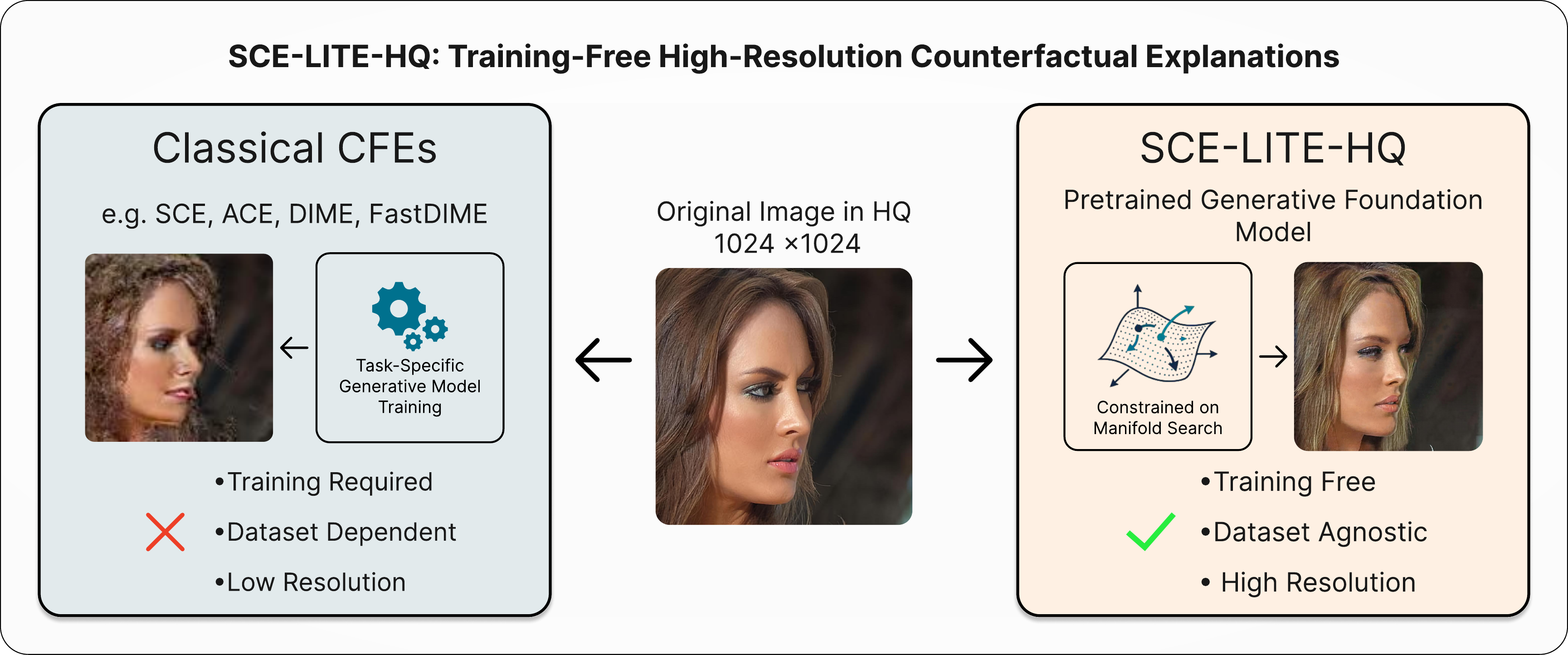}
    \caption{Comparison between SCE-LITE-HQ and traditional counterfactual explanation frameworks. Conventional methods (left) rely on computationally intensive training of dataset-specific generative models, whereas SCE-LITE-HQ (right) leverages pretrained Generative Foundation Models (GFMs). We illustrate counterfactuals generated on the CelebA-HQ dataset for the blond attribute. SCE-LITE-HQ produces counterfactuals at the original image resolution while preserving visual quality. In contrast, FastDIME (left) requires adjusting the image resolution, which leads to lower-quality counterfactuals.}
    \label{fig:method_overview}
\end{figure*}

\section{Introduction}
\label{sec:intro}
Deep neural networks have become indispensable tools in computer vision  \cite{CHAI2021100134}, achieving state-of-the-art performance across a wide range of tasks. Despite their predictive power, these models often operate as "black boxes," providing users with little insight into the underlying reasoning behind their decisions. This lack of transparency is particularly problematic in high-stakes applications where human trust and accountability are paramount. Furthermore, these models are prone to exploiting spurious correlations to reach decisions - a phenomenon known as the "Clever Hans" effect. Consequently, the field of Explainable Artificial Intelligence (XAI)  \cite{Samek_2021} has emerged to develop methods that illuminate the inner workings of these complex models.

Among these methods, Counterfactual Explanations (CFEs)  \cite{10.1145/3677119} offer a minimalist and intuitive approach to interpretability. A CFE identifies the smallest possible modifications to an input required to flip a model’s prediction to a target class. Beyond mere visualization, CFEs are grounded in causal reasoning; they provide users with actionable insights by demonstrating not just where a model focused its attention, but exactly what features must change to alter the outcome. This makes CFEs a promising diagnostic tool for identifying "shortcut features"  \cite{Geirhos_2020}. By revealing whether a model relies on irrelevant artifacts rather than robust, class-defining characteristics, giving rise to so-called Clever Hans predictors  \cite{Lapuschkin_2019}. On the other hand, CFEs enable us to mitigate these biases and improve model robustness, as demonstrated in frameworks like CFKD \cite{bender2023towards} \cite{bender2025mitigatingcleverhansstrategies}.

Despite the utility of counterfactual reasoning, a robust CFE framework must navigate a complex set of competing constraints. To be practically effective, a method must generate counterfactuals that satisfy several key desiderata  \cite{BENDER2026112811}: they must be realistic, remaining firmly on the natural data manifold; sparse, introducing only the minimal modifications necessary to flip the classification; diverse, providing multiple plausible paths to the target outcome; and we argue that it should also be scalable, maintaining computational efficiency even in high-dimensional spaces. Achieving these goals simultaneously remains a significant challenge in computer vision. 

Formally, for a classifier $f: \mathcal{X} \rightarrow \mathbb{R}^c$ and an input $x \in \mathcal{X}$, a counterfactual explanation involves finding a minimally changed input $x' = x + \delta x$ that causes the model to produce a different, predetermined output. And so we can formulate the counterfactual explanation problem as follows:
\begin{equation}
    x' = \argmin_{x' \in \mathcal{X}} \, d(x, x') \quad \text{subject to} \quad f(x') = y' \text{ and } x' \in \mathcal{M}
    \label{eq:count_opt}
\end{equation}

Where $d(\cdot, \cdot)$ represents a metric defined on the input space $\mathcal{X}$ quantifying the magnitude of the perturbation required to transform the original instance $x$ into the counterfactual instance $x'$, $f$ is the model we want to explain, $y'$ is our new target value,and $\mathcal{M}$ is the data manifold.

Generating high-quality visual counterfactuals remains a significant challenge. Images are inherently high-dimensional, making it difficult to identify meaningful directions for modification that can transform a factual instance into a counterfactual one. Optimization in pixel space is particularly problematic, as small perturbations are rarely semantically meaningful and tend to produce adversarial or unrealistic artifacts  \cite{https://doi.org/10.1002/widm.1567}  \cite{10646794}. To address this, recent methods employ generative models such as GANs \cite{10.1145/3422622}, normalizing flows \cite{10.5555/3045118.3045281}, or diffusion models \cite{10.5555/3495724.3496298} to better constrain counterfactuals to the data manifold  \cite{7348689} and ensure realism. While these approaches improve the perceptual quality of generated counterfactuals, they typically rely on dataset-specific generative models that must be trained for each domain or task. This dependency makes them computationally expensive and limits their scalability, especially for high-resolution images. 
 
In this work, we leverage generative foundation models (GFMs) to address these limitations. GFMs are pretrained on massive and diverse datasets, learning rich and structured latent spaces that capture broad visual distributions \cite{latent_space}. They have demonstrated strong zero-shot generation capabilities across a wide range of domains and image resolutions, suggesting that they can serve as scalable backbones for counterfactual explanation. 
We introduce SCE-LITE-HQ, a scalable framework that leverages GFMs for high-resolution counterfactual generation. Our contributions are threefold:
\begin{itemize}
    \item Dataset-agnostic generation: We operate directly in the latent space of pretrained GFMs, eliminating task-specific training.
    \item Stable optimization: We integrate smoothed gradient guidance and mask-based diversification to ensure robustness and realism.
    \item High-resolution scalability: We demonstrate valid, diverse counterfactuals at resolutions up to $1024\times 1024$ previously unattainable by existing methods.
\end{itemize}

Experiments across natural and medical image datasets show that SCE-LITE-HQ matches or exceeds state-of-the-art methods in validity, fidelity, sparsity, and diversity, while significantly reducing computational overhead.

\section{Related Work}
\label{sec:related}

While there are works to counterfactuals in other domains such as tabular data~\cite{mothilal2020explaining}, graphs~\cite{bechtoldt2025graph}, proteins~\cite{klosprotein} or natural language~\cite{sarkar2024explaining} this section reviews prior work on counterfactual explanations for visual models  \cite{9710816} \cite{ha2025diffusion}  \cite{10.1007/978-3-031-19775-8_23} \cite{10.1007/978-3-031-26293-7_14}  \cite{10205255}  \cite{10345703} \cite{bender2026visual}, with a focus on generative approaches and their scalability limitations, and discusses recent efforts to leverage foundation models for explainability. We position our method relative to both lines of work, highlighting how it addresses key shortcomings in resolution, efficiency, and generality.

A variety of approaches have been proposed  \cite{10345703}  \cite{10205255}  \cite{weng2024fast}  \cite{10.1007/978-3-031-26293-7_14}  \cite{10205255}  \cite{10484315} for producing counterfactuals of image classifiers, each aiming to overcome challenges with the optimization problem itself and the quality of the resulting counterfactuals. Early methods relied on generative models to constrain edits to the data manifold, though they often struggled with output diversity and fidelity.

To better ensure manifold adherence, Diffeomorphic Counterfactuals (DiffeoCF)~ \cite{10345703} and DiVE ~ \cite{9710816} utilize invertible latent space models, allowing for exact latent space inversions and smooth transformations. However, these methods are limited by the expressiveness of the invertible model and do not scale well to high-resolution images.

The rise of diffusion models \cite{10.5555/3495724.3496298} significantly advanced the quality of generated counterfactuals  \cite{10.1007/978-3-031-26293-7_14}. and gave rise to multiple methods that leverage their generative power. These methods can be categorized by how they integrate classifier guidance and update the image:

For instance, ACE (Adversarial Visual Counterfactual Explanations)  \cite{10205255} 
utilizes classifier gradients filtered through a diffusion model's denoising process to steer clear of 
non-robust adversarial directions. This method combines $L_1$ and $L_2$ regularization to maintain 
proximity to the original image and employs RePaint  \cite{9880056} 
as a post-processing step to encourage feature sparsity. Similarly, DIME (Diffusion Models 
Counterfactual Explanations)  \cite{10.1007/978-3-031-26293-7_14}, and its 
successor, FastDIME  \cite{weng2024fast}, optimize within 
the latent space; however, they differ from ACE by encoding the factual image into a latent state only 
once at the beginning of the process, subsequently updating that state via classifier guidance. 
FastDIME further improves efficiency by introducing a gradient approximation and mask-based 
inpainting, which successfully reduces computational complexity from $O(T^2)$ to $O(T)$. 


The Smooth Counterfactual Explorer (SCE)~ \cite{BENDER2026112811} integrates mechanisms from these prior works, while introducing novel components: smooth distilled surrogates to stabilize gradients and lock-based diversifiers to ensure sufficiency and diversity. However, SCE and all previously mentioned methods share a critical limitation: they require training or fine-tuning a dataset-specific generative model, which restricts their scalability and generalizability, particularly to high-resolution settings.


\textbf{Foundation Models (FMs)}  \cite{10834497}, pretrained on vast and diverse datasets, have revolutionized machine learning by providing powerful, general-purpose representations. Their application in Explainable AI (XAI) is an emerging frontier. For instance, TIME~ \cite{10484315} demonstrates the potential of using Stable Diffusion for counterfactual explanations by learning and manipulating textual biases. However, TIME and similar approaches often require complex, multi-stage fine-tuning of the foundation model for the specific explanation task, which limits flexibility and scalability. In contrast, we argue that a more direct approach is to leverage the rich, pretrained latent spaces of these models without such task-specific fine-tuning. This promising direction bypasses costly retraining and naturally scales to generate high-resolution, realistic counterfactuals across diverse domains.

\textbf{Positioning of Our Work}. Our work bridges the gaps identified above by leveraging the rich, pretrained latent spaces of generative foundation models (GFMs) without any task-specific fine-tuning. 



    
    

\section{Background: From Diffusion to Rectified Flow}
\label{sec:background}

Our method leverages the latent space of generative foundation models, most of which are based on diffusion \cite{10.5555/3495724.3496298} or rectified flow principles \cite{liu2023flow} as a realistic image prior to constrain counterfactual search to the data manifold. We briefly review these foundations, highlighting why rectified flow is particularly advantageous for scalable, high-resolution counterfactual optimization.

\subsection{Diffusion Models}
Denoising Diffusion Probabilistic models (DDPM) \cite{10.5555/3495724.3496298} are latent variable generative models that learn to reverse a gradual noising process applied to data. The forward diffusion is a fixed Markov chain that adds Gaussian noise to data $\mathbf{x}_0 \sim q(\mathbf{x}_0)$ over $T$ steps:
\begin{equation}
q(\mathbf{x}_t | \mathbf{x}_{t-1}) = \mathcal{N}(\mathbf{x}_t; \sqrt{1 - \beta_t} \mathbf{x}_{t-1}, \beta_t \mathbf{I}),
\end{equation}
with variance schedule ${\beta_t} \in (0,1)_{t=1}^{T}$.The marginal distribution of $\mathbf{x}_t$ conditioned on $\mathbf{x}_0$ admits a closed form:
\begin{equation}
q(\mathbf{x}_t | \mathbf{x}_0) = \mathcal{N}(\mathbf{x}_t; \sqrt{\bar{\alpha}_t} \mathbf{x}_0, (1 - \bar{\alpha}_t)\mathbf{I}),
\end{equation}
where $\bar{\alpha}_t = \prod_{s=1}^{t} (\alpha_s)$. and $\alpha_s = 1-\beta_s$ Reparameterizing with standard noise $\boldsymbol{\epsilon} \sim \mathcal{N}(\mathbf{0}, \mathbf{I})$ yields
\begin{equation}
\mathbf{x}_t = \sqrt{\bar{\alpha}_t} \mathbf{x}_0 + \sqrt{1 - \bar{\alpha}_t} \boldsymbol{\epsilon},
\end{equation}
which expresses $\mathbf{x}_t$ as a linear mixture of signal and noise.

The reverse process learns a denoising model $p_\theta(\mathbf{x}_{t-1} | \mathbf{x}_t)$ that reconstructs data from noise, starting from $p(\mathbf{x}_T) = \mathcal{N}(\mathbf{0}, \mathbf{I})$.

\subsection{Rectified Flow: A Deterministic Alternative}
Rectified Flow (RF)  \cite{liu2023flow} offers a deterministic, ODE-based alternative that is better suited for stable latent optimization. It learns a continuous velocity field $\mathbf{v}_\theta(\mathbf{x}_t, t)$ that transports samples from a source distribution $p_1$ (noise) to a target distribution $p_0$ (data) along straight paths:
\begin{equation}
\mathbf{x}_t = (1-t)\mathbf{x}_0 + t \mathbf{\epsilon}, \quad t \in [0,1],
\end{equation}
where $\mathbf{\epsilon} \sim p_1$ and $\mathbf{x}_0 \sim p_0$. The model is trained via Conditional Flow Matching (CFM):
\begin{equation}
\min_\theta \int_0^1 \mathbb{E} \left[ \| \mathbf{v}_\theta(\mathbf{x}_t, t) - (\mathbf{\epsilon} - \mathbf{x}_0) \|^2 \right] dt.
\end{equation}
Once trained, generation follows the ODE:
\begin{equation}
\frac{d\mathbf{x}_t}{dt} = \mathbf{v}_\theta(\mathbf{x}_t, t), \quad \mathbf{x}_1 \sim p_1,
\label{eq:ode_flow}
\end{equation}
yielding samples $\mathbf{x}_0 \sim p_0$ without stochastic sampling.

\vspace{0.5em}
\noindent

Unlike stochastic diffusion, RF defines a deterministic, straight-line trajectory between noise and data  \cite{liu2023flow}.
This reduces variance during gradient-based optimization, enabling more stable and efficient high-resolution counterfactual generation.


\section{SCE-LITE-HQ}
\label{sec:method}

\begin{figure*}[t!]
    \centering
    \includegraphics[width=0.9\linewidth]{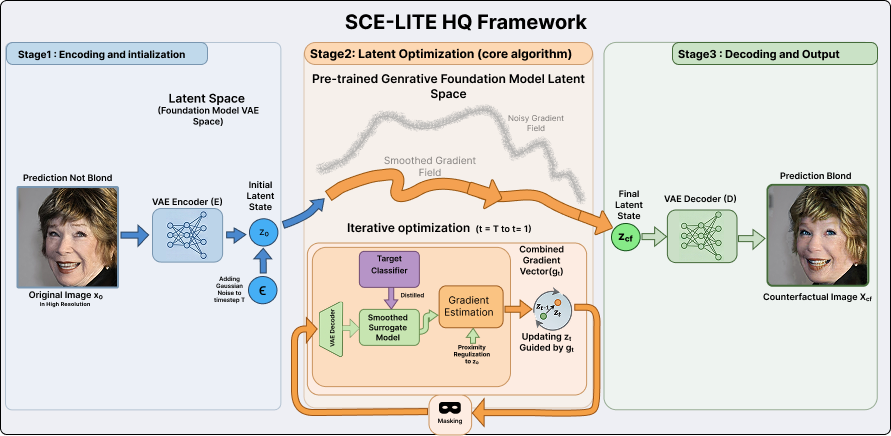} 
    \caption{Overview of the SCE-LITE-HQ algorithm for generating counterfactual images via latent-space optimization guided by a foundation model and classifier gradients.}
    \label{fig:method_overview1}
\end{figure*}

SCE-LITE-HQ is a scalable framework for generating high-quality visual counterfactual explanations that overcomes two key limitations of prior work: (1) dependency on dataset-specific generative models, and (2) restriction to low resolution. Our method builds upon insights from Smooth Counterfactual Explorer (SCE)~ \cite{BENDER2026112811}, specifically its smoothed surrogate guidance and diversification principles, and adapts the efficient gradient approximation of FastDIME~ \cite{weng2024fast} to operate within the latent space of pretrained foundation models.

\paragraph{Latent Space Foundation.} We operate within the compressed latent space $\mathcal{Z}$ of pretrained foundation models. Let $\mathcal{E}$, $\mathcal{D}$ denote the encoder/decoder. For image $\mathbf{x}_0$, its latent is $\mathbf{z}_0 = \mathcal{E}(\mathbf{x}_0) \in \mathbb{R}^{c\times h\times w}$. This representation enables efficient high-resolution optimization while preserving semantics.

Concretely, our counterfactual search is achieved by minimizing the following loss:
\begin{equation}
    \mathcal{L} = \underbrace{\beta \ell(\hat{f}(\mathcal{D}(\hat{\mathbf{z}}_t)), y')}_{\text{classifier guidance}} + \underbrace{\lambda_1\|\mathbf{z}_t - \mathbf{z}_0\|_1 + \lambda_2\|\mathbf{z}_t - \mathbf{z}_0\|_2}_{\text{proximity loss}}
\end{equation} within the generative process of the foundation model that acts as a prior constraining our search on the data manifold, where the individual components are detailed in the following subsections.

\subsection{Smoothed Surrogate Guidance}
\label{subsec:surrogate}
Directly utilizing gradients from the original classifier $f$ often results in noisy optimization trajectories and the generation of non-robust adversarial artifacts. Following SCE~ \cite{BENDER2026112811}, we employ a knowledge distillation \cite{hinton2015distillingknowledgeneuralnetwork} framework to train a smoothed surrogate classifier $\hat{f}$. This surrogate is designed to approximate the decision boundaries of $f$ while yielding a more stable and smoother gradient field.

To enhance the regularity of the loss landscape, the surrogate is trained via a multi-objective optimization:
\begin{equation}
\begin{aligned}
\min_{\hat{f}} \mathbb{E}_{(\mathbf{x},y)} \left[ \mathcal{L}_{\text{KL}}(\hat{f}(\mathbf{x}) \parallel f(\mathbf{x})) + \lambda_1 \mathcal{L}_{\text{MixUp}} + \lambda_2 \mathcal{L}_{\text{LS}} + \lambda_3 \mathcal{L}_{\text{Adv}} \right]
\end{aligned}
\end{equation}
where $\mathcal{L}_{\text{KL}}$ denotes the Kullback-Leibler divergence, $\mathcal{L}_{\text{MixUp}}$ \cite{zhang2018mixup} to promote local linearity  $\mathcal{L}_{\text{LS}}$ label smoothing  \cite{10.5555/3454287.3454709} to penalize overconfidence, and $\mathcal{L}_{\text{Adv}}$ adversarial training \cite{ijcai2021p591} to enforce gradient stability. Architecturally, $\hat{f}$ mirrors $f$ but substitutes standard activations with \text{LeakyReLU} and \text{Softplus}. For a detailed motivation and theoretical analysis of these regularization terms, check ~\cref {app:surrogate_smoothness}.
We define the surrogate guidance signal $\mathbf{g}_{\text{smooth}}(\mathbf{z})$ as the gradient with respect to a latent $\mathbf{z}$:
\begin{equation}
\mathbf{g}_{\text{smooth}}(\mathbf{z}) = \nabla_{\mathbf{z}} \ell(\hat{f}(\mathcal{D}(\mathbf{z})), y')
\label{eq:surr_grad}
\end{equation}
where $\ell$ is the classification loss (e.g cross-entropy) with respect to target class $y'$, and $\mathcal{D}$ is the decoder mapping latent $z$ to image space.
This term serves as the primary directional signal for all subsequent counterfactual search operations.
By optimizing against $\hat{f}$, we effectively bypass the high-frequency noise of the original model. As demonstrated in ~\cref{sec:ablation_gradient_source}, this significantly improves the fidelity and robustness of the resulting counterfactuals.

\subsection{Generative Latent Trajectories}
\label{subsec:optimization}

Following ~\cref{sec:background},but now operating on  the compressed latent representation $\mathcal{Z}$ rather than the pixel space $\mathcal{X}$. We consider counterfactual generation as guidance applied during the reverse-time latent generation of a pretrained foundation model. Rather than directly perturbing pixels or a single latent code, SCE-LITE-HQ operates on the latent trajectory induced by the generative model, enabling counterfactual edits that remain consistent with the learned data manifold.

For a broad class of generative transport models, intermediate latent states can be written as
\begin{equation}
\mathbf{z}_t = a_t \mathbf{z}_0 + b_t \boldsymbol{\epsilon}, \quad \boldsymbol{\epsilon} \sim \mathcal{N}(0, \mathbf{I}),
\end{equation}

In rectified flow (RF) models, $a_t = 1 - t$ and $b_t = t$, and reverse-time generation follows a deterministic ordinary differential equation (ODE) (~\cref{eq:ode_flow}). In diffusion models, $a_t = \sqrt{\bar{\alpha_t}}$ and $b_t = \sqrt{1 - \bar{\alpha}_t}$, with reverse-time generation governed by a stochastic differential equation.

This unified view characterizes the generative process, which is later perturbed by classifier-derived guidance, forming the basis for efficient counterfactual optimization described next.

\subsection{Efficient Gradient Approximation}
\label{subsec:gradient}
Computing gradients through the full iterative transport chain is computationally expensive.
We utilize FastDIME's  \cite{weng2024fast} efficient gradient approximation, following the architectural flow shown in ~\cref{fig:method_overview1}

At each optimization step $t$, we compute a one-step denoised estimate of $z_0$ from the current noisy state $\mathbf{z}_t$.

For diffusion-based foundation models, the denoised estimate is given by
\begin{equation}
\hat{\mathbf{z}}_t =
\frac{\mathbf{z}_t - \sqrt{1 - \bar{\alpha_t}}\,\epsilon_\theta(\mathbf{z}_t, t)}
{\sqrt{\bar{\alpha_t}}},
\end{equation}
where $\epsilon_\theta$ denotes the predicted noise.

For rectified flow models, we instead compute
\begin{equation}
\hat{\mathbf{z}}_t = \mathbf{z}_t - t \cdot \mathbf{v}_\theta(\mathbf{z}_t, t),
\end{equation}

The total updated gradient $g_t$ is then constructed by combining the classifier guidance and proximity penalties 
\begin{equation}
\begin{aligned}
    \mathbf{g}_t = &\  \beta \cdot \mathbf{g}_{\text{smooth}}(\hat{\mathbf{z}}_t) + \nabla_{\mathbf{z}_t} \left( \lambda_1 \|\mathbf{z}_t - \mathbf{z}_0\|_1 + \lambda_2 \|\mathbf{z}_t - \mathbf{z}_0\|_2 \right)
\end{aligned}
\end{equation}

where $\beta$ and $\lambda_{1,2}$ balance the objectives. The latent state is updated as:
\[
\mathbf{z}_t \leftarrow \mathbf{z}_t - \eta \cdot \mathbf{g}_t
\]

This formulation ensures a functional separation of objectives: the classification signal is derived from the 'clean' semantic estimate $\hat{\mathbf{z}}_t$, while the proximity penalties $\mathcal{L}_{prox}$ act directly on the optimization variable $\mathbf{z}_t$ to anchor the trajectory to the manifold of the factual image

\subsection{Mask-Based Diversification}
\label{subsec:diversification}
While the L1/L2 proximity penalties globally discourage large deviations, they cannot enforce spatially localized edits. To address this, we implement a dual-phase masking strategy that ensures sparsity by restricting edits to the minimal spatial region necessary for the class flip, and diversity by forcing subsequent counterfactuals to explore disjoint semantic regions.

\textbf{Phase I: Adaptive Masking for Sparsity.} 
To achieve sparse edits, we dynamically constrain the optimization to regions where the current denoised estimate $\hat{\mathbf{x}}_0^{(t)}$ significantly deviates from the original image $\mathbf{x}_0$. For each denoising step, we compute a binary mask based on the spatial residual:
\begin{equation}
    \mathbf{m}_t = \mathbb{I} \left[ \Psi \left( \mathbf{x}_0 , \hat{\mathbf{x}}_0^{(t)}\right) < \tau_{\text{inpaint}} \right]
\end{equation}
where $\Psi(\cdot)$ denotes the pixel-wise absolute difference smoothed by a Gaussian kernel. The generated mask is then projected into the foundation  model latent space and then apply inpainting:
\begin{equation}
    \mathbf{z}_t \leftarrow \mathbf{z}_t \odot (1 - \downarrow \mathbf{m}_t) + \mathbf{z}_{\text{0}} \odot \downarrow \mathbf{m}_t
\end{equation}
where $\downarrow$ represents the downsampling operation to the latent resolution. This mechanism restricts the flow to "inpaint" only the most salient features necessary for class flip.

\textbf{Phase II: Cumulative Exclusion for Diversity.} 
To prevent the convergence to the same visual explanation across multiple runs, we utilize a diversity constraint for the $k$-th counterfactual ($k > 1$). We maintain a cumulative difference map of the previous counterfactuals:
\begin{equation}
    \mathbf{C}^{(k)} = \sum_{j=1}^{k-1} \Psi \left( \mathbf{x}_{\text{cf}}^{(j)} , \mathbf{x}_0 \right)
\end{equation}
The exclusion mask $\mathbf{m}_{\text{excl}}^{(k)} = \mathbb{I}[\mathbf{C}^{(k)} < \tau_{\text{inpaint}}]$ is held fixed throughout the $k$-th generation. By setting $\mathbf{m}_{\text{fixed}}^{(k)} = \mathbf{m}_{\text{excl}}^{(k)}$, we effectively "block" previously identified features, forcing the flow to find alternative causal paths for the target prediction.

\section{Experimental Setup}
\label{sec:experiments}

\textbf{Datasets and tasks.} We evaluate SCE-LITE-HQ across different visual domains and resolutions. For natural images, we use CelebA~ \cite{liu2015faceattributes} at $128\times128$ resolution for attribute classification (Blond Hair, Smiling). We also evaluate on CelebA-HQ~ \cite{CelebAMask-HQ} at $1024\times1024$ resolution to assess high-resolution scalability. To evaluate the effectiveness of generated counterfactuals for model debiasing, we construct a poisoned variant of CelebA-Blond where we introduce a spurious correlation between hair color and gender (blond hair predominantly female, dark hair predominantly male), simulating scenarios where a classifier might rely on gender as a decision boundary rather than hair color. For medical imaging, we use a subset of Camelyon17~ \cite{10.1093/gigascience/giy065} with an introduced spurious correlation between histopathological patches and their hospital of origin (malignant samples from one hospital, benign from another).

\textbf{Models.} We employ ResNet-18 ~\cite{7780459} classifiers fine-tuned per dataset and task as our predictive models. Unlike prior methods requiring task-specific generative model training, SCE-LITE-HQ leverages pretrained foundation models: Stable Diffusion 3~ \cite{DBLP:conf/icml/EsserKBEMSLLSBP24} for natural images and PathLDM~ \cite{Yellapragada_2024_WACV} for medical images.

\textbf{Baselines.} We benchmark against ACE  \cite{10205255}, DIME \cite{10.1007/978-3-031-26293-7_14}, FastDIME  \cite{weng2024fast}, and SCE  \cite{BENDER2026112811}, with all baselines reimplemented and hyperparameters optimized for fair comparison.

\textbf{Evaluation Metrics.} We assess counterfactual quality across multiple desiderata  \cite{BENDER2026112811}:
\begin{itemize}
    \item \textbf{Sufficiency (NAFR):} Non-Adversarial Flip Rate (NAFR) measures the flip rate of a smoothed classifier $\hat{f}$ that is trained similarly to our method's surrogate classifier ~\cref{subsec:surrogate}.
    \item \textbf{Sufficiency (Diversity):} For each sample, we generate two counterfactuals and compute \(1 - \text{CosSim}(\Delta_1, \Delta_2)\), where \(\Delta_i = e(\mathbf{x}) - e(\mathbf{x}_{\text{cf}}^{(i)})\) and \(e\) is an encoding function.

    \item \textbf{Understandability (Sparsity):} sparsity of edits given by \(1 - \text{avg}(|\Delta|) / \max(|\Delta|)\) with \(\Delta = e(\mathbf{x}) - e(\mathbf{x}_{\text{cf}})\).
    \item \textbf{Fidelity (NA):} Non-adversarial rate.
\begin{equation}
\resizebox{0.98\linewidth}{!}{$\displaystyle
\text{NA} = \frac{1}{N_{\text{flip}}}\sum_{i=1}^N \mathbb{I}\left(f(\mathbf{x}_{\text{cf}}^{(i)}) \neq f(\mathbf{x}^{(i)}) \land \hat{f}(\mathbf{x}_{\text{cf}}^{(i)}) \neq \hat{f}(\mathbf{x}^{(i)})\right)
$} 
\end{equation} where $f$ is the original classifer, $\hat{f}$ smoothed surrogate
    \item \textbf{Actionability (Gain):} In-the-Loop Gain after applying CFKD  \cite{bender2025mitigatingcleverhansstrategies} Gain = \((\text{Acc}_{\text{after}} - \text{Acc}_{\text{before}}) / \text{Err}_{\text{before}}\).
\end{itemize}

\section{Results}

~\cref{tab:results_main} presents the quantitative evaluation across all datasets and metrics. SCE-LITE-HQ achieves competitive or superior performance across all desiderata without dataset-specific generative model training.

\textbf{Fidelity and sufficiency.} Across all tasks, SCE-LITE-HQ consistently achieves top-tier fidelity, closely matching or surpassing task-specific models on most benchmarks. On clean datasets, SCE slightly outperforms SCE-LITE-HQ; we attribute that to the use of a task-specific diffusion model, but SCE-LITE-HQ remains highly competitive using off-the-shelf foundation models. On poisoned or spurious datasets, SCE-LITE-HQ stands out, achieving the highest fidelity and the strongest downstream gains, highlighting its robustness to spurious correlations.
Regarding sufficiency metrics, SCE-LITE-HQ achieves competitive scores across all tasks. In settings with strong spurious correlations, reduced sufficiency reflects the model’s reliance on a small set of dominant features

\begin{figure*}[t!]
    \centering
    \includegraphics[width=0.9\linewidth]{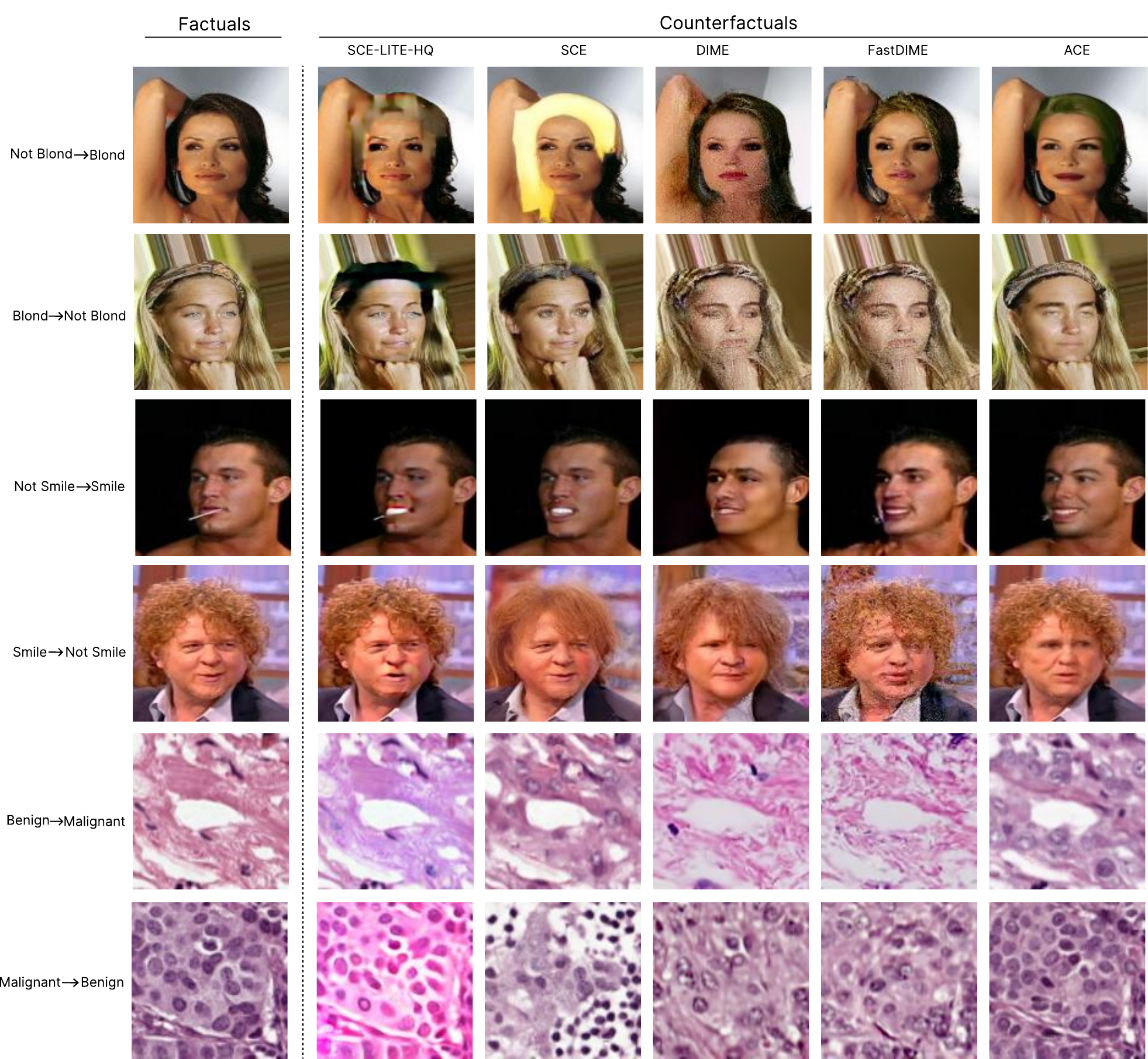}
    \caption{Comparison of counterfactual explanations generated by different methods across multiple tasks and datasets. For each example, we show the original input, its predicted label, and the counterfactual generated by each method. On the blond hair classification task, SCE-LITE-HQ achieves a favorable balance between Sparsity, realism, and robustness, successfully flipping the model's prediction without excessive perturbations (in contrast to SCE, which overshoots, or other methods, which undershoot). On the Camelyon dataset, SCE-LITE-HQ identifies confounding factors, enabling the generation of counterfactuals that expose spurious correlations; these can be used with the CFKD algorithm to fine-tune the original model and improve accuracy, explaining the substantial performance gains achieved by SCE-LITE-HQ on this dataset.}
    \label{fig:compare_methods}
\end{figure*}

\textbf{Understandability.} SCE-LITE-HQ consistently achieves high sparsity across all tasks, indicating sparse and semantically targeted edits. ~\cref{fig:compare_methods} shows that SCE-LITE-HQ modifies fewer attributes than baseline methods while maintaining perceptual quality comparable to task-specific generative models, despite operating at a lower resolution of (128), well below the recommended resolutions for Stable Diffusion 3 (1024) and PathLDM (256).

\textbf{Actionability.} SCE-LITE-HQ demonstrates strong debiasing performance. On poisoned CelebA-Blond Male, it achieves 21.6\% gain, matching DIME. On Camelyon17, SCE-LITE-HQ achieves 47.9\% gain exceeding second best baseline by a factor of 2, illustrating the benefit of leveraging foundation models trained on diverse data rather than potentially biased task-specific generators.

\begin{figure}[htbp]
    \centering
    \includegraphics[width=0.95\linewidth]{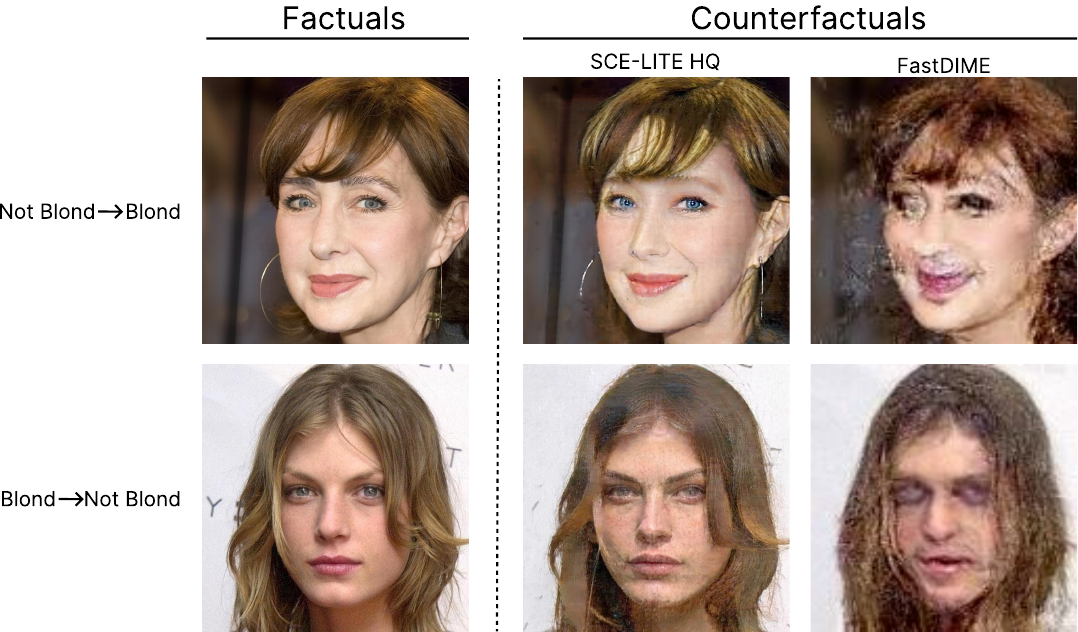}
    \caption{Visualization of counterfactual explanations generated at high resolution on the CelebA-HQ dataset using SCE-LITE-HQ and FastDIME as a baseline. SCE-LITE-HQ successfully identifies the confounding attribute of blue eyes, which is highly correlated with blond hair, while FastDIME fails to capture this relationship due to information loss at lower resolutions.}
    \label{fig:hq_generated}
\end{figure}

\textbf{Resolution scalability.} ~\cref{fig:hq_generated} shows SCE-LITE-HQ scales to 1024 $\times$ 1024 without retraining, identifying subtle confounders, eye color, and hair color, which would not be possible to identify in low resolution.

\textbf{Limitations.} SCE-LITE-HQ's NA rate lags SCE on some tasks (e.g., CelebA-Blond), suggesting task-specific generators still offer a slight advantage when computational budget allows. 

Overall, SCE-LITE-HQ delivers a strong balance of explanation quality, interpretability, and practical effectiveness across clean and spurious datasets, while eliminating the need for extensive generative model training. These results support foundation models as a robust and scalable alternative for counterfactual explanations.

\begin{table*}[t!]
\centering \footnotesize
\renewcommand{\arraystretch}{1.1}
\setlength{\tabcolsep}{4pt}
\begin{tabular}{llrcccccccc}
\toprule
 & & & \multicolumn{4}{c}{\cellcolor{CadetBlue!20}\textbf{desiderata}}\\
 & & & \multicolumn{1}{c}{\cellcolor{CadetBlue!20}{fidelity}} & \multicolumn{1}{c}{\cellcolor{CadetBlue!20}{understandability}} & \multicolumn{2}{c}{\cellcolor{CadetBlue!20}{sufficiency}} & \\
Dataset / Model & Method & & \cellcolor{CadetBlue!20}(NA) & \cellcolor{CadetBlue!20}(sparsity) & \cellcolor{CadetBlue!20}(NAFR) & \cellcolor{CadetBlue!20}(diversity) & gain \\ \midrule
\multirow{5}{*}{Celeb-A Blond}
 & ACE & & 73.3 & \underline{71.7}  & 44.0 & 47.3 & -\\
 & DIME & & 64.4 & 71.1 & 58.0 & 51.7 & -\\
 & FastDIME & & 62.5 & 71.3 & 50.0 & 39.9 & - \\
 & SCE & & \textbf{100.0} & 71.4 & \textbf{90.0} & \textbf{74.6} & -\\
 \rowcolor{gray!10}\cellcolor{white}
 & \textbf{SCE-LITE-HQ (ours)} & & \underline{86.7} & \textbf{72.1} & \underline{72.0} & \underline{62.4} & - \\
\midrule
\multirow{5}{*}{Celeb-A Smile}
 & ACE & & \underline{98.0} & 71.0 & \textbf{96.0} & 19.3 & -\\
 & DIME & & 90.0 & \underline{72.0} & \underline{90.0} & 43.5 & -\\
 & FastDIME & & 78.0 & 71.6 & 78.0 & 39.2 & -\\
 & SCE & & 92.1 & 71.3 & 70.0 & \textbf{74.2} & -\\
 \rowcolor{gray!10}\cellcolor{white}
 & \textbf{SCE-LITE-HQ (ours)} & & \textbf{100.0} & \textbf{72.6}& 71.0 & \underline{50.8} & - \\
\midrule
\multirow{5}{*}{CelebA-Blond (Male)}
 & ACE & & 38.8 & \textbf{72.3} & 67.5 & 43.2 & 13.4\\
 & DIME & & 19.48 & 70.06 & 19.48 & 38.48 & \textbf{21.7}\\
 & FastDIME & & 68.46 & 70.88 & 20.0 & \underline{42.56} & 13.6 \\
 & SCE & & \underline{98.86} & \underline{71.8} & \textbf{86.5} & \textbf{45.6} & 15.01\\
 \rowcolor{gray!10}\cellcolor{white}
 & \textbf{SCE-LITE-HQ (ours)} & & \textbf{100.0} & 71.6 & \underline{82.8} & 35.8 & \underline{21.6} \\
\midrule
\multirow{5}{*}{Camelyon17 (Hosp.)}
 & ACE & & 32.8 & 40.9 & 19.0 & 5.7 & 7.0\\
 & DIME & & 47.8 & \textbf{65.8} & 44.0 & 12.5 & -3.8\\
 & FastDIME & & 36.0& 48.0 & 27.0 & 3.5 & 16.3 \\
 & SCE & & \textbf{95.3} & 37.6 & \textbf{50.5} & \underline{32.2} & \underline{22.2}\\
 \rowcolor{gray!10}\cellcolor{white}
 & \textbf{SCE-LITE-HQ (ours)} & & \underline{75.8} & \underline{53.9} & \underline{47.5} & \textbf{38.9} & \textbf{47.9} \\
\bottomrule
\end{tabular}
\caption{Quantitative evaluation of counterfactual explanation methods across desiderata metrics. We evaluate fidelity (NA: Non-Adversarial rate), understandability (sparsity in latent space), and sufficiency (NAFR: Non-Adversarial Flip Rate, diversity). The gain column measures improvement in model accuracy after fine-tuning with CFKD \cite{bender2025mitigatingcleverhansstrategies} algorithm. SCE-LITE-HQ achieves competitive performance on standard benchmarks (CelebA-Blond, CelebA-Smile) while demonstrating superior robustness on the poisoned dataset (CelebA-Blond Male) and substantial gains on medical imaging (Camelyon17), effectively identifying spurious correlations for model debiasing. Bold indicates best performance; underline indicates second-best.}
\label{tab:results_main}
\end{table*}

\section{Ablation Study: Effect of Gradient Source}

A core component of SCE-LITE-HQ is the use of a smooth surrogate classifier to refine the gradient field used for counterfactual optimization. Traditional counterfactual explanation methods rely on vanilla gradients from the original classifier, which are often noisy and poorly aligned with semantic features  \cite{10.5555/3305381.3305417}, frequently resulting in adversarial or implausible counterfactuals. In this ablation, we isolate the effect of the gradient source by comparing gradients obtained from our distilled classifier against vanilla gradients and several attribution-based alternatives. Specifically, we evaluate whether gradient smoothing or attribution-based methods can serve as effective substitutes for raw classifier gradients, including SmoothGrad  \cite{smilkov2017smoothgradremovingnoiseadding}, SmoothDiff  \cite{hill2025smoothed}, and Integrated Gradients (IG)  \cite{10.5555/3305890.3306024}.

\label{sec:ablation_gradient_source}
\begin{table}[t!]
\centering
\caption{Effect of gradient source on counterfactual generation (CelebA-Blond, ResNet-18).}
\label{tab:ablation_gradient_source}
\resizebox{\columnwidth}{!}{
\begin{tabular}{lcccc}
\toprule
\textbf{Gradient Source} & \textbf{NAFR} & \textbf{NA} & \textbf{Sparsity} & \textbf{Diversity} \\
\midrule
Vanilla Gradient & 28.0 & 44.4 & 73.2 & \textbf{75.5} \\
SmoothGrad & 56.0 & 80.0 & 72.8 & 61.3 \\
SmoothDiff & 35.0 & 52.2 & \textbf{74.1}& 75.0 \\
Integrated Gradients (IG) & 36.0 & 55.4 & 73.2 & 74.6 \\
SmoothGrad $\times$IG & 18.0 & 29.5 & 70.8 & 67.3 \\
SmoothDiff $\times$IG & 23.0 & 34.3 & 70.4 & 70.1 \\
\midrule
\textbf{Smoothed Classifier (Ours)} & \textbf{72.0} & \textbf{86.7} & 72.1 & 62.4 \\
\bottomrule
\end{tabular}
}
\end{table}

As shown in ~\cref{tab:ablation_gradient_source}, replacing vanilla gradients with those from our distilled classifier yields a substantial improvement in the Non-Adversarial Rate (NA), rising from 44.4\% to 86.7\% on the CelebA-Blond task, while maintaining competitive sparsity and diversity. Smoothing techniques like SmoothGrad and SmoothDiff improve upon the baseline, they fail to match the performance of our smoothed model. Furthermore, while attribution methods improve local smoothness, they are not designed as optimization objectives and remain suboptimal for counterfactual generation. These results demonstrate that the limitations of prior methods stem largely from the quality of the gradient signal; by utilizing distilled, semantically aligned gradients, SCE-LITE-HQ achieves superior robustness and fidelity. The exploratory inclusion of attribution-based gradients, while not yet optimal, opens a promising direction for bridging attribution methods and counterfactual explanation techniques in future work.

\section{Conclusion}

We introduced \textbf{SCE-LITE-HQ}, a scalable framework for visual counterfactual explanations that leverages pretrained generative foundation models without task-specific fine-tuning. By combining latent-space optimization in foundation models and a two-phase masking strategy for sparsity and diversity, SCE-LITE-HQ enables counterfactual generation at resolutions up to 1024$\times$1024, previously unattainable in prior counterfactual pipelines.

Experimental results across natural and medical domains show that SCE-LITE-HQ matches or exceeds state-of-the-art methods in validity, fidelity, sparsity, and diversity, while eliminating the computational overhead of per-dataset generator training. Ablation studies further indicate that isolating high-quality gradient guidance rather than relying solely on smoother optimization landscapes is essential for producing semantically faithful counterfactuals.

More broadly, SCE-LITE-HQ highlights a shift toward adapting foundation models for explainable AI, using them as reusable components within counterfactual explanation pipelines. This perspective opens the door to scalable, high-resolution explanations in domains where trust and interpretability are critical.

\clearpage
\bibliography{egbib}
\bibliographystyle{icml2026}
\clearpage
\appendix
\crefalias{section}{appendixsection}
\section{Surrogate Smoothness Regularization: Motivation and Analysis}
\label{app:surrogate_smoothness}

This appendix provides a detailed motivation and theoretical analysis for the regularization terms 
and architectural modifications employed in training the smoothed surrogate classifier $\hat{f}$ 
(~\cref{subsec:surrogate}). The overarching goal is to construct a surrogate whose loss landscape 
admits a well-behaved gradient field,  one that mitigates the shattered gradient 
problem~\cite{10.5555/3305381.3305417} and ensures that gradient-based 
Counterfactual search yields semantically meaningful image edits rather than degenerate adversarial artifacts.

\paragraph{MixUp ($\mathcal{L}_{\text{MixUp}}$).}
MixUp~\cite{zhang2018mixup} enforces linear behavior between training examples by augmenting data with convex combinations:
\begin{equation}
    \tilde{\mathbf{x}} = \alpha \mathbf{x}_i + (1 - \alpha) \mathbf{x}_j, \quad \tilde{y} = \alpha y_i + (1 - \alpha) y_j,
\end{equation}
where $\alpha \sim \text{Beta}(\beta, \beta)$ and $(\mathbf{x}_i, y_i), (\mathbf{x}_j, y_j)$ are randomly sampled training pairs. The MixUp loss is:
\begin{equation}
    \mathcal{L}_{\text{MixUp}} = \mathbb{E}_{\alpha, i, j} \left[ \mathcal{L}_{\text{CE}}(\hat{f}(\tilde{\mathbf{x}}), \tilde{y}) \right],
\end{equation}
where $\mathcal{L}_{\text{CE}}$ denotes cross-entropy loss. Mathematically, MixUp encourages the surrogate to satisfy:
\begin{equation}
    \hat{f}(\alpha \mathbf{x}_i + (1-\alpha) \mathbf{x}_j) \approx \alpha \hat{f}(\mathbf{x}_i) + (1-\alpha) \hat{f}(\mathbf{x}_j),
\end{equation}
promoting local linearity. This directly reduces the Hessian norm $\|\nabla^2 \ell(\hat{f}(\mathbf{x}), y)\|$, which in turn bounds gradient variation and decreases $L_{\nabla}$.

\paragraph{Label Smoothing ($\mathcal{L}_{\text{LS}}$).}
Label smoothing~\cite{10.5555/3454287.3454709} replaces hard one-hot targets $y$ with softened distributions:
\begin{equation}
    y_c^{\text{smooth}} = (1 - \epsilon) y_c + \frac{\epsilon}{C},
\end{equation}
where $\epsilon \in [0,1]$ controls smoothing strength. The loss becomes:
\begin{equation}
    \mathcal{L}_{\text{LS}} = \mathbb{E}_{(\mathbf{x}, y)} \left[ \mathcal{L}_{\text{CE}}(\hat{f}(\mathbf{x}), y^{\text{smooth}}) \right].
\end{equation}
This prevents the model from becoming overconfident at decision boundaries, effectively penalizing large logit magnitudes. Since $\nabla_{\mathbf{x}} \ell(\mathbf{x}, y) = \nabla_{\mathbf{x}} f(\mathbf{x}) \cdot \nabla_{f} \ell(f, y)$ (where $f = \hat{f}(\mathbf{x})$ are logits), reducing logit magnitudes directly dampens gradient magnitudes, leading to smoother optimization trajectories.

\paragraph{Adversarial Robustness ($\mathcal{L}_{\text{Adv}}$).}
We incorporate adversarial training~\cite{ijcai2021p591} to explicitly penalize gradient sensitivity. For each training sample $(\mathbf{x}, y)$, we generate adversarial perturbations:
\begin{equation}
    \mathbf{x}^{\text{adv}} = \mathbf{x} + \delta, \quad \text{where} \quad \delta = \arg\max_{\|\delta\|_p \leq \epsilon_{\text{adv}}} \mathcal{L}_{\text{CE}}(\hat{f}(\mathbf{x} + \delta), y).
\end{equation}
The adversarial loss is:
\begin{equation}
    \mathcal{L}_{\text{Adv}} = \mathbb{E}_{(\mathbf{x}, y)} \left[ \mathcal{L}_{\text{CE}}(\hat{f}(\mathbf{x}^{\text{adv}}), y) \right].
\end{equation}
Adversarial training enforces gradient stability by requiring consistent predictions under small perturbations. This can be viewed as minimizing:
\begin{equation}
    \max_{\|\delta\|_p \leq \epsilon_{\text{adv}}} \left\| \nabla_{\mathbf{x}} \ell(\hat{f}(\mathbf{x}), y) - \nabla_{\mathbf{x} + \delta} \ell(\hat{f}(\mathbf{x} + \delta), y) \right\|,
\end{equation}
which directly constrains the gradient Lipschitz constant $L_{\nabla}$ in a local neighborhood around training examples.

\paragraph{Architectural Modifications for Smoothness}
Beyond regularization, we modify the architecture of $\hat{f}$ to promote gradient smoothness. Standard activations like ReLU introduce non-differentiability at zero, causing gradient discontinuities. We replace these with smooth alternatives:
\begin{itemize}
    \item \textbf{LeakyReLU:} $f(z) = \max(\alpha z, z)$ with small $\alpha > 0$, ensuring non-zero gradients for negative inputs.
    \item \textbf{Softplus:} $f(z) = \log(1 + e^z)$, providing $C^{\infty}$ smoothness with $f'(z) = \sigma(z)$ (sigmoid).
\end{itemize}
These substitutions ensure that $\nabla_{\mathbf{z}} \hat{f}(\mathbf{z})$ is continuous everywhere, eliminating gradient discontinuities inherent in ReLU-based networks.

\section{Algorithm Details}
\label{appendix:algorithm}
We provide the pseudocode for the SCE-LITE-HQ optimization loop in ~\cref{alg:sce_lite}. This implementation corresponds to the theoretical framework discussed in ~\cref{sec:method}, specifically highlighting the integration of the foundation model's velocity field and the surrogate-based guidance.

\begin{algorithm}[h!] 
\caption{SCE-LITE-HQ: Scalable Counterfactual Explanations with Foundation Models}
\label{alg:sce_lite}
\begin{algorithmic}[1] 
\renewcommand{\algorithmicrequire}{\textbf{Input:}}
\renewcommand{\algorithmicensure}{\textbf{Output:}}

\REQUIRE $\mathbf{x}_0$: Input image, $y^*$: Target class, $f$: Classifier, $\mathcal{G}_\theta$: Foundation model, $\mathcal{E},\mathcal{D}$: VAE encoder/decoder, $T$: Time step, $\eta$: Learning rate, $\beta,\lambda$: Weights
\ENSURE $\mathbf{x}_{\text{cf}}$: Counterfactual image

\STATE \COMMENT{\textit{Step 1: Train smoothed surrogate}}
\STATE $\hat{f} \leftarrow \text{TrainSurrogate}(f)$

\STATE \COMMENT{\textit{Step 2: Encode to latent space}}
\STATE $\mathbf{z}_0 \leftarrow \mathcal{E}(\mathbf{x}_0)$
\STATE $\boldsymbol{\epsilon} \sim \mathcal{N}(\mathbf{0}, \mathbf{I})$
\STATE $\mathbf{z}_t \leftarrow (1-t)\mathbf{z}_0 + t\boldsymbol{\epsilon}$ for $t=T$

\STATE \COMMENT{\textit{Step 3: Iterative optimization}}
\FOR{$t = T, T-1, \dots, 1$}
    \STATE \COMMENT{\textit{Predict velocity field}}
    \STATE $\mathbf{v}_\theta \leftarrow \mathcal{G}_\theta(\mathbf{z}_t, t)$
    \STATE \COMMENT{\textit{ODE integration}}
    \STATE $\mathbf{z}_t \leftarrow \mathbf{z}_t + \Delta t \cdot \mathbf{v}_\theta$
    \STATE \COMMENT{\textit{Compute denoised estimate}}
    \STATE $\hat{\mathbf{z}}_t \leftarrow \mathbf{z}_t - t \cdot (\mathbf{v}_\theta)$
    \STATE \COMMENT{\textit{Compute gradients}}
    \STATE $\mathbf{g}_t \leftarrow \beta \nabla \ell_c + \lambda \nabla \ell_d$
    \STATE \COMMENT{\textit{Update latent}}
    \STATE $\mathbf{z}_t \leftarrow \mathbf{z}_t - \eta \cdot \mathbf{g}_t$
    \STATE \COMMENT{\textit{Apply mask}}
    \STATE $\mathbf{z}_t \leftarrow (1-\mathbf{m}_t)\mathbf{z}_t + \mathbf{m}_t\mathbf{z_0}$
\ENDFOR

\STATE \COMMENT{\textit{Step 4: Decode to pixel space}}
\STATE $\mathbf{x}_{\text{cf}} \leftarrow \mathcal{D}(\mathbf{z}_t)$
\RETURN $\mathbf{x}_{\text{cf}}$
\end{algorithmic}
\end{algorithm}

\section{Computational Cost}
\label{sec:appendix_compute}

We benchmark SCE-LITE-HQ against baseline methods (ACE, DIME, FastDIME, SCE) in terms of the computational cost of counterfactual generation on the CelebA-Blond task. All results are averaged over three independent runs, each generating counterfactuals for 40 samples.

\textbf{Hardware and software.} Experiments were conducted on a single NVIDIA H100 PCIe GPU using CUDA~12.1, PyTorch~2.1.2, and Python~3.9.

 SCE-LITE-HQ achieves the lowest per-sample latency (0.66 s) while requiring substantially less peak GPU memory than ACE and SCE.

These results demonstrate that SCE-LITE-HQ offers substantial efficiency and scalability advantages, making it suitable for large-scale and iterative debiasing settings where both speed and memory efficiency are crucial.

\begin{table}[H]
\centering
\caption{Runtime and memory consumption of counterfactual generation methods on poisoned CelebA-Blond Male (40 samples, 3 runs average, NVIDIA H100 GPU). Peak GPU Mem. measures maximum memory during computation.}
\label{tab:cf_runtime_memory}
\resizebox{\columnwidth}{!}{
\begin{tabular}{lrrrrr}
\toprule
\textbf{Method} & \textbf{Total Time (s)} & \textbf{Time/Sample (s)} & \textbf{Peak GPU (MB)} & \textbf{Peak CPU (MB)} \\
\midrule
ACE        & 1001.99 & 25.05 & 40961 & 2803 \\
DIME       & 1081.61 & 27.04 & 965  & 2867 \\
FastDIME   & 70.29   & 1.76  & 964  & 2983 \\
SCE        & 1295.81 & 32.40 & 52999  & 3046 \\
\midrule
\textbf{SCE-LITE-HQ} & 26.24 & 0.66 & 11569 & 8468 \\
\bottomrule
\end{tabular}
}
\end{table}
\label{app:qualitative}


\begin{figure*}[t]
    \centering
    \includegraphics[width=\linewidth]{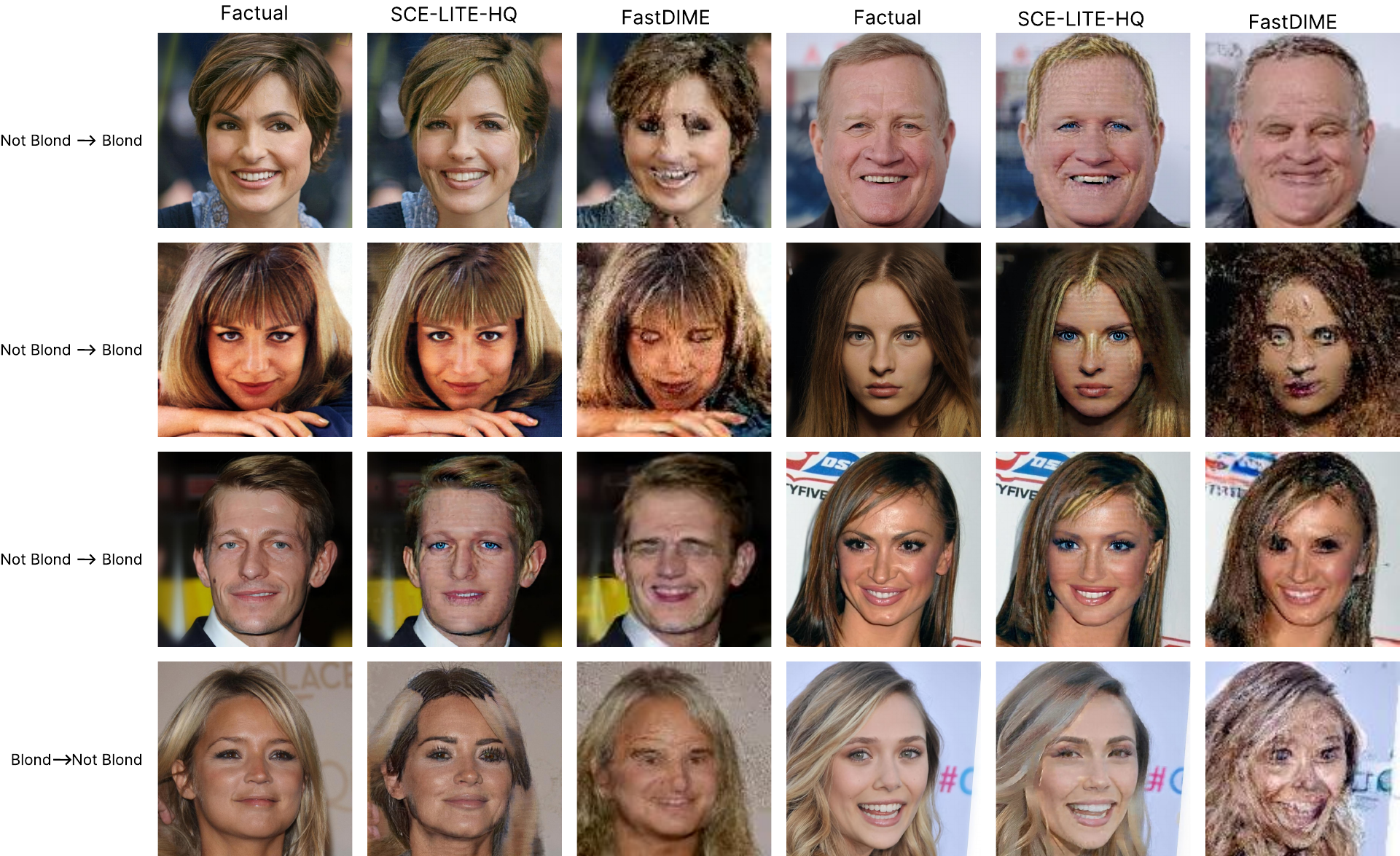}
    \caption{High-resolution counterfactual examples on CelebA-HQ. We compare SCE-LITE-HQ against FastDIME, illustrating qualitative differences in visual fidelity and semantic coherence.}
    \label{fig:qualitative_appendix1}
\end{figure*}

\begin{figure*}[t]
    \centering
    \includegraphics[width=\linewidth]{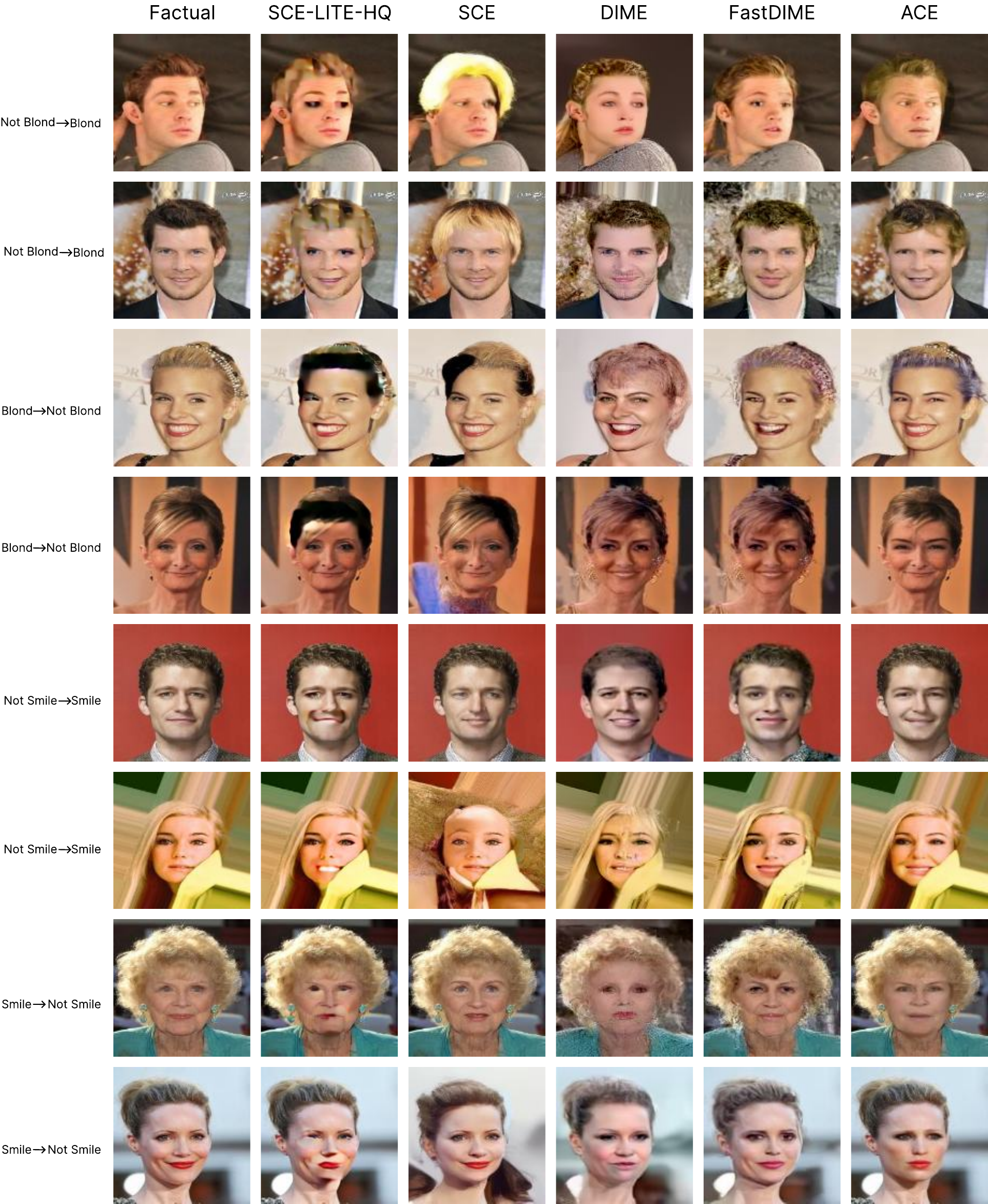}
    \caption{Additional qualitative comparison of counterfactual explanations. SCE-LITE-HQ consistently identifies the key attributes for flipping predictions while avoiding unnecessary modifications, demonstrating improved sparsity and semantic relevance over prior methods.}
    \label{fig:qualitative_appendix2}
\end{figure*}

\begin{figure*}[t]
    \centering
    \includegraphics[width=\linewidth]{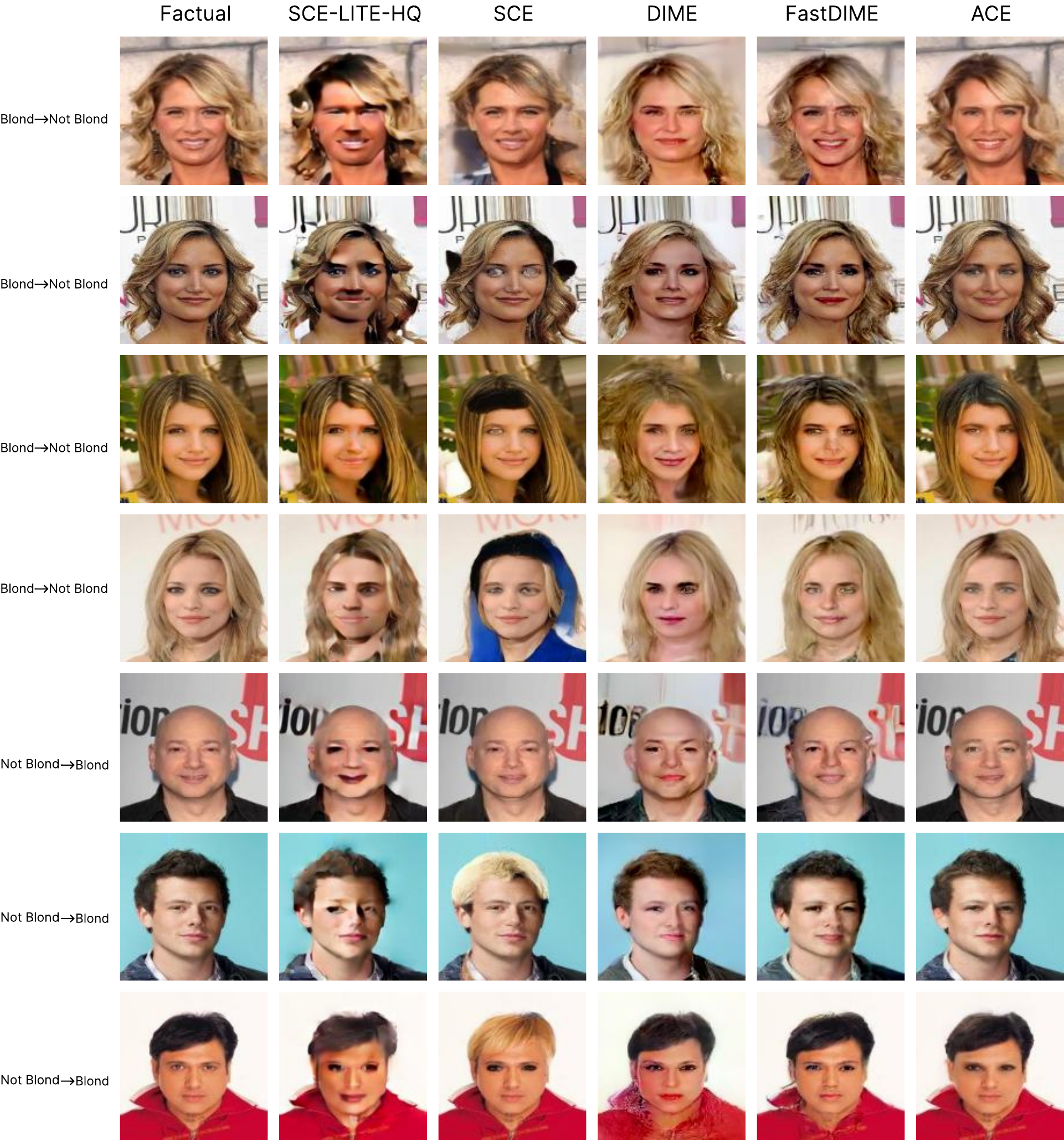}
    \caption{Qualitative Comparison on Poisoned CelebA. SCE-LITE-HQ correctly identifies the underlying bias: it transforms blond female factuals into male-featured counterfactuals to satisfy the classification flip. Conversely, for dark-haired males, the model introduces feminine semantic features while preserving the original hair color. These counterfactuals can later be used to debias the model using the CFKD algorithm \cite{bender2025mitigatingcleverhansstrategies}.}
    \label{fig:qualitative_appendix3}
\end{figure*}

\end{document}